\documentclass[10pt,a4paper,conference,composocconf]{IEEEtran}
\usepackage{cite}
\ifCLASSINFOpdf
   \usepackage[pdftex]{graphicx}
  \graphicspath{{img/}}
  \DeclareGraphicsExtensions{.pdf,.jpeg,.png}
\else
\fi
\usepackage{amsmath}
\usepackage{array}
\usepackage{url}

\usepackage[utf8]{inputenc}

\usepackage{amssymb} 
\usepackage{mathtools}
\usepackage{bm}
\usepackage{xspace}
\usepackage{microtype}
\usepackage{multicol}
\usepackage{booktabs}

\usepackage{tikz}
\usepackage{pgfplots}
\usetikzlibrary{patterns,external,pgfplots.groupplots,fit,calc,positioning,shapes}
\usepackage{booktabs}

\usepackage{varioref}
\usepackage[pagebackref=true,breaklinks=true,letterpaper=true,colorlinks,bookmarks=false,citecolor=green!80!black,linkcolor=red!80!black,urlcolor=blue]{hyperref}
\usepackage{cleveref}
\usepackage{subcaption}
\usepackage{csquotes}
\usepackage{siunitx}

\newcommand{\onedot}{.\xspace}
\newcommand{\etal}[1]{#1~et~al\onedot}

\newcommand{\cf}{cf\onedot}
\newcommand{\ie}{i.\,e.,\xspace}

\newcommand{\aka}{a.\,k.\,a\onedot}

\renewcommand{\vec}[1]{\bm{#1}}
\DeclareMathOperator*{\argmin}{argmin}
\crefname{section}{Sec.}{Sections}
\crefname{figure}{Fig.}{Figure}
\crefname{table}{Tab.}{Table}
\crefname{equation}{Equ.}{Equation}

\definecolor{faublue}{RGB}{0,51,102}

\newcommand{\bx}{\vec{x}}
\newcommand{\bmu}{\vec{\mu}}
\newcommand{\map}{mAP\xspace}
\newcommand{\icdard}{ICDAR13\xspace}
\newcommand{\khatt}{KHATT\xspace}
\newcommand{\cvl}{CVL\xspace}
\newcommand{\iam}{IAM\xspace}

\newcommand{\ltwo}{\ell_2}
\DeclareMathOperator{\sign}{sign}
\NewDocumentCommand{\rot}{O{60} O{1em} m}{\makebox[#2][l]{\rotatebox{#1}{#3}}}%
\NewDocumentCommand{\rotn}{O{90} O{1em} m}{\makebox[#2][l]{\rotatebox{#1}{#3}}}%
\NewDocumentCommand{\rotninety}{O{90} O{1em} m}{\makebox[#2][l]{\rotatebox{#1}{#3}}}%

\begin{document}
\title{Encoding CNN Activations for Writer Recognition}

\author{\IEEEauthorblockN{Vincent Christlein, Andreas Maier}
\IEEEauthorblockA{
Pattern Recognition Lab, Friedrich-Alexander-Universität Erlangen-Nürnberg,
 91058 Erlangen, Germany\\
 vincent.christlein@fau.de, andreas.maier@fau.de
}}

\maketitle

\begin{abstract}
The encoding of local features is an 
essential part for writer identification and writer retrieval. 
While CNN activations have already been used as local features in related works,
the encoding of these features has attracted little attention so far. 
In this work, we compare the established VLAD encoding with triangulation
embedding. We further investigate generalized max pooling as an alternative to
sum pooling and  
the impact of decorrelation and Exemplar SVMs. 
With these techniques, we set new standards on two publicly available datasets (\icdard, \khatt).
\end{abstract}


\IEEEpeerreviewmaketitle

\section{Introduction}
Handwritings play an important role for law enforcement agencies in proving
someone's authenticity because it can be used as a biometric identifier like
faces or speech. Forensic experts are consulted to make a decision in such
scenarios. However, searching for a particular writer in a huge dataset requires
automatic or semi-automatic methods. Due to the mass-digitization processes of
historical documents, this topic has also attracted attention in the field of
historical document analysis~\cite{Brink12WIU,Gilliam10SIM,Fecker14WIF}.

In this work, we focus on the task of offline writer recognition, in particular
\emph{writer identification} and \emph{writer retrieval}. Writer identification
denotes the problem of finding the writer of a query handwriting in a dataset of
known writers. For writer retrieval, all handwritings of a
dataset are ranked according to their similarity to the query sample.

Writer identification\,/\,retrieval methods can be grouped into  
\emph{codebook}-based methods and \emph{codebook-free} methods.
Codebook-based methods rely on a codebook that serves as background model. This
model is used to compute statistics that form the global descriptor which is
compared with each other, a process known as \emph{encoding}. 
Codebook-free methods compute a global image descriptor directly from the
handwriting. For example, the width of the ink trace~\cite{Brink12WIU} or the
angles of stroke directions~\cite{He14DHR} were used for writer identification
purposes. 

We employ features extracted from a deep
convolutional neural network (CNN). CNNs are the state-of-the-art tool for image
classification since the AlexNet CNN~\cite{Krizhevsky12} won the ImageNet
competition. They also became more and more popular in the domain of document
analysis~\cite{Bluche13FEC,Jaderberg14DFT,Wahlberg16ICFHR} and recently also in
the field of writer
recognition~\cite{Fiel15CAIP,Christlein15GCPR,Xing16,Tang16}.

In this work, we investigate multiple parts of a codebook-based writer
identification\,/\,retrieval pipeline using features computed by means of a CNN.
We try to answer the following questions:
1) Produces a deeper network better activations?
2) Which encoding method performs better? More specifically, we compare
vectors of locally aggregated descriptors (VLAD)~\cite{Jegou12ALI} with the more recent triangulation
embedding~\cite{Jegou14TEA}, which showed superior performance to other
encoding methods using traditional features~\cite{Murray16} and CNN activation features~\cite{Hoang17} 
3) Another recent improvement concerns the aggregation of local descriptors. Does generalized max pooling (GMP)~\cite{Murray14} work better than sum pooling?
4) Which effects do PCA whitening on CNN activations have and how much impact do Exemplar SVMs have?

The paper is organized as follows: After reporting about related work in
\cref{sec:related_work}, the general pipeline, the encoding methods and
generalized max pooling are outlined in \cref{sec:methodology}.
In \cref{sec:evaluation}, the different parts of our pipeline are evaluated
before the paper is concluded in \cref{sec:conclusion}. 

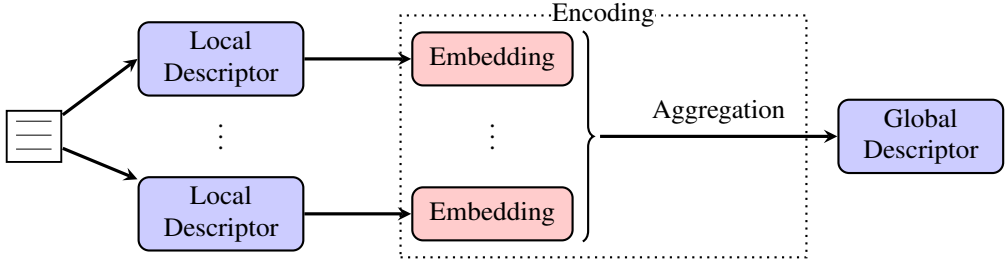
\begin{figure*}[t]
	\centering
	\tikzstyle{block} = [rectangle, draw, fill=blue!20, 
		text width=5.5em, text centered, rounded corners, 
	minimum height=2em, thick]
	\tikzstyle{roundish} = [draw, rectangle, fill=red!20, rounded corners,
	text width=6em, text centered, minimum height=2em, thick, inner sep=0em]
	\tikzstyle{hhilit} = [draw=black, thick, dotted, inner xsep=4pt, 
	inner	ysep=6pt]
	\tikzstyle{newpart}=[align=center,black,thin,draw, rounded	corners,
	fill=green!20]
	\tikzsetnextfilename{pipeline}
	\begin{tikzpicture}[>=stealth]
		\node[block] (l1) {Local Descriptor};
		\node[block, above=3em of l1] (l2) {Local Descriptor};
		\node[inner sep=0pt,xshift=-7em] at  ($(l1)!0.5!(l2)$) (doc) {\includegraphics[height=2em]{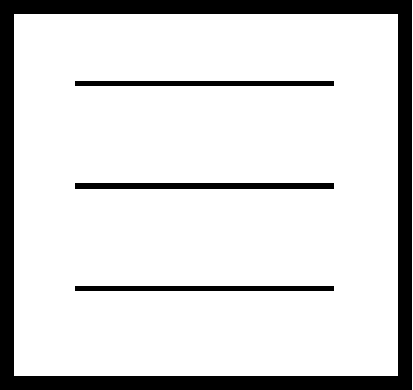}};
		\node[roundish, right=4em of l1] (s1) {Embedding};
		\node[roundish, right=4em of l2] (s2) {Embedding};
		\node[block, right=29em of doc] (sv) {Global Descriptor};
		\node[yshift=0.1cm](d1) at ($(l1)!0.5!(l2)$){$\vdots$};
		\node[yshift=0.1cm] (d1) at ($(s1)!0.5!(s2)$){$\vdots$};
		\draw[->,very thick] (doc) -- (l1);
		\draw[->,very thick] (doc) -- (l2.west);
		\draw[->,very thick] (l1) -- (s1);
		\draw[->,very thick] (l2) -- (s2);
		\draw [decorate,decoration={brace,amplitude=4pt,raise=2pt},thick]
		($(s2.north east) + (0.05cm,0)$) -- ($(s1.south east) + (0.05cm,0)$)
		node[midway,align=center,sloped,above=1pt](b){};
		\draw[->,very thick](b) -- (sv) node[midway,sloped,above=1pt](agg){Aggregation};
		\node[hhilit,fit=(s1) (s2) (agg)] (upperdots) {}; 
		\node[fill=white, inner sep=0pt] at ($(upperdots.north)$) (title1) {Encoding};
	\end{tikzpicture}
	\caption{Encoding of local descriptors to form a global representation which
	can be compared.}
	\label{fig:pipeline}
\end{figure*}

\section{Related Work}
\label{sec:related_work}
One of the first writer recognition methods using deep learning techniques was
introduced by Fiel and Sablatnig~\cite{Fiel15CAIP}. 
They proposed the use of convolutional neural
network (CNN) activations
obtained from the penultimate layer of a trained CNN. 
The eight layer deep CNN was trained by 
word or line segmentations obtained from the \iam dataset.
The mean of the CNN activation features were used as feature
vector which were compared using the $\chi^2$-distance. They improved upon the
state of the art (at that time) on the \iam and the ICFHR'12 datasets. At the
same time, they show a worse result on the \icdard dataset (containing Greek
and English samples), which can be attributed to wrong word segmentations and
the missing Greek training data.

Instead, we proposed to encode the CNN activation features by means of GMM
supervectors~\cite{Christlein15GCPR}, which are subsequently compared by the
cosine distance. We showed improved \map on the \icdard, \cvl and \khatt
dataset.

In contrast to these works~\cite{Fiel15CAIP,Christlein15GCPR}, Xing and
Quiao~\cite{Xing16} proposed the use of a network structure consisting of two
branches sharing the convolutional layers. Two adjacent image patches of the
same writer, \ie two non-overlapping windows of an input line, are used as
input for the network. The authors report good performance results on the
\iam dataset. 
However, the CNN is trained for a specific writer on a
line-basis using some lines for training, one for validation and one for testing.
In other words, they used an end-to-end training for the writers which makes a
comparison with other retrieval\,/\,identification-based publications impossible.

Tang and Wu~\cite{Tang16} considered the global appearance as an important
feature. They do not employ a patch-wise approach but use full document images
which are created artificially by means of random segmented words from the
original writing. In this way, they generated about \num{500} training and
about twenty testing samples per writer to get more diverse feature vectors. 
The similarity is computed by means of the log-likelihood ratio of the CNN
activation features of the penultimate layer. They presented the current best
results on the \cvl dataset and the second best results using the \icdard
dataset.

\section{Methodology}
\label{sec:methodology}
Commonly, encoding local descriptors is achieved by computing statistics of a
background model, such as $k$-means, to embed each descriptor into a higher
dimensional space, which are then aggregated. A popular encoding technique is
VLAD~\cite{Jegou12ALI}, which aggregates first order statistics with sum
pooling.  The global representation is typically compared using the cosine
distance. 

In this work we investigate how VLAD encoding of CNN activations compares to the more recent
triangulation embedding~\cite{Jegou14TEA}. 
We further explore generalized max pooling~\cite{Murray14}, which balances the
embeddings with regard to the final similarity score. 
To the best of our knowledge this has not yet been explored for CNN activations. 
Additionally, we evaluate the effect of PCA whitening on the
different parts of the pipeline. Instead of using the cosine distance, we
further explore exemplar classifiers to learn a similarity between
representations.

\subsection{Feature Extraction}

Similar to our previous work~\cite{Christlein15GCPR}, we use the activations
from the penultimate layer from a trained CNN as features. The CNN is trained
with four million raw (i.\,e.\ non-normalized) $32\times32$ patches randomly
sampled from the script contour. The contours are obtained by a connected
component analysis of the binarized images. We evaluate different network
topologies to investigate their influence on the retrieval performance. The CNN
activation features are then encoded to a global representation. 

\subsection{Encoding}
Encoding consists of two steps, \cf \cref{fig:pipeline}: i) an embedding step
where a, possible non-linear, function is applied to the local feature vectors
in order to create a high dimensional representation, and ii) an aggregation
step where the embedded local descriptors are pooled into a fixed-length vector.
After a possible normalization step, the global representations can be compared
to each other.  In case of image retrieval this is commonly done by means of the
cosine distance between the global representations.

\subsubsection{Aggregation}
Sum pooling is the standard aggregation method for pooling local descriptors. 
Max pooling requires embedding
functions that associate a certain strength to one visual word which is not directly applicable
to the most common embedding techniques that rely on higher order statistics~\cite{Murray16}. 
The drawback of sum pooling is that it assumes that the descriptors in an image are
independently and identically distributed (iid). More frequently occurring
descriptors will be more influential in the final representation, \aka \emph{visual
burstiness}. Thus,
encoding methods have to normalize for this. In general, this is achieved
during embedding or normalization. Alternatively, \emph{generalized max pooling}
(GMP)~\cite{Murray14} and \emph{democratic aggregation}~\cite{Jegou14TEA} were
proposed to counter this in the aggregation step. Both methods re-weight the patch statistics such that
their influence regarding the final similarity score is equalized. Comparing
both approaches~\cite{Murray16}, GMP showed superior performance. As indicated
by its name, it generalizes max pooling by formulating the balancing problem of
any embedding $\phi(\bx) \in \mathbb{R}^D, \bx \in \mathbb{R}^{D_l}$ as ridge regression
problem. Therefore, let~\cite{Murray16}
\begin{equation}
	\label{eq:gmp}
	\phi(\bx)^{\top}\vec{\xi}_{\text{gmp}}(\mathcal{X}) = C, \quad \forall\bx\in\mathcal{X}\,,
\end{equation}
where $\bx$ is a local descriptor of the set of all local image descriptors
$\mathcal{X}$, $\vec{\xi}_{\text{gmp}}$ is the GMP representation and $C$ is a
constant which can be set arbitrarily since it has no influence when the
representation is eventually $\ltwo$-normalized. 
\Cref{eq:gmp} can be generalized for the $D\times n$ matrix $\vec{\Phi}$ of all $n$ patch
embeddings to
\begin{equation}
	\label{eq:gmp2}
	\vec{\Phi}^{\top}\vec{\xi}_{\text{gmp}} = \vec{1}_n, 
\end{equation}
where $\vec{1}_n$ denotes the vector of all constants set to $1$. 
This linear system can be turned into a least-squares ridge regression problem:
\begin{equation}
	\label{eq:ridge}
	\vec{\xi}_{\text{gmp}} = \argmin_{\vec{\xi}} \lVert
	\vec{\Phi}^{\top}\vec{\xi} - \vec{1}_n \rVert^2 +
	\lambda\lVert\vec{\xi}\rvert^2\;,
\end{equation}
with $\lambda$ being a
regularization term that stabilizes the solution.
\Cref{eq:ridge} can be computed using conjugate gradient descent (CGD). 
In the remainder, $\vec{\psi}$ denotes the aggregated result. CGD can typically be
applied component-wise, \ie on each $\vec{\phi}_k$ (see below), to speed up the process. 

\subsubsection{VLAD Embedding}
VLAD can be viewed as a
non-probabilistic version of the Fisher Kernel~\cite{Jegou12ALI} encoding only
first order statistics. In combination with
improvements such as whitening~\cite{Jegou12NEA},
intra-normalization~\cite{Arandjelovic13AAV}, or residual
normalization~\cite{Delhumeau13}, VLAD is one of the standard encoding techniques.
We successfully employed it for writer identification~\cite{Christlein15ICDAR}
and it has already been used in combination with deep-learning-based features for
the task of classification and retrieval~\cite{Gong14MSO,Ng15,Paulin16}. 

For VLAD encoding, the embedding function $\vec{\phi}$ for the local descriptor
$\bx$ is computed by the 
residual to its nearest cluster center $\bmu_k$ of the dictionary
$\mathcal{D}=\{\bmu_k\in\mathbb{R}^{D_l}, k=1,\ldots,K\}$:
\begin{align}
\label{eq:vlad}
\vec{\phi}_{\text{VLAD},k}(\bx)
	&= \alpha_{k}(\bx) (\bx - \bmu_k) \;\\
 \alpha_{k}(\bx) &= 
	\begin{dcases}
		1&\text{if}~ k=\argmin_{j=1,\dotsc,K} \lVert \bx - \bmu_j \rVert_2\\
		0&\text{else}
	\end{dcases}
	\;.
\end{align}
The full embedding follows as
$\vec{\phi}_{\text{VLAD}}=(\vec{\phi}^{\top}_1,\ldots,\vec{\phi}^{\top}_K)^{\top}$.

VLAD encodings can be normalized to counter visual burstiness.
The most popular normalization technique is 
\emph{power normalization}, where each element $\psi_i$ of the aggregated embedding
$\vec{\psi}$
is normalized as: 
\begin{equation}
	\hat{\psi}_i = \sign(\psi_i) \lvert \psi_i \rvert^{p}\,,\forall \psi_i \in
	\vec{\psi} \;,
\end{equation}
where $p$ is typically chosen to be $0.5$. This is also known as signed square root (SSR),
which resembles the Hellinger kernel. Power normalization is typically followed by
an $\ltwo$ normalization. 

Other proposed normalization techniques are \emph{residual
normalization}~\cite{Delhumeau13} and \emph{intra
normalization}~\cite{Arandjelovic13AAV}.  In the former one, each embedding
$\vec{\phi}_k$ is $\ltwo$-normalized. In intra normalization, the component-wise
aggregated embeddings $\vec{\psi}_k$ are $\ltwo$ normalized.
It is also beneficial to decorrelate and whiten the representations to prevent
an over-counting of co-occurrences~\cite{Jegou12NEA}. This is commonly achieved
by means of a principal component analysis with whitening (PCA whitening). 
PCA whitening can be applied either
at the global representation $\vec{\psi}$ or at the embeddings $\vec{\phi}$ as
well as their individual components $\vec{\psi}_k$ and $\vec{\phi}_k$, where the
latter one was suggested by \etal{Delhumeau}~\cite{Delhumeau13} and referred to
\emph{local coordinate systems} (LCS). 

\subsubsection{Triangulation Embedding}
Triangulation embedding~\cite{Jegou14TEA} is very similar to VLAD with
residual normalization. The normalized residuals of anchor points (= cluster
centers) to feature descriptors are computed. These can be viewed as directions
discarding absolute distances to cluster centers. The key difference is that no
association function ($\alpha$) is applied to the
vectors. All normalized residuals are encoded, not just the nearest neighbor:
\begin{equation}
	\label{eq:tri}
	\vec{\phi}_{\text{T-Emb},k}(\bx_t) = \frac{\bx_t - \bmu_k}{\lVert \bx_t - \bmu_k \rVert_2} \;.
\end{equation}
Each embedding is whitened using PCA.
However, instead of a dimensionality reduction, the authors propose to discard
the \emph{largest} $D_l$ eigenvalues and corresponding eigenvectors.  This has
the effect of reducing the variance of the cosine similarity between unrelated
descriptors~\cite{Jegou14TEA}. Each embedding is subsequently $\ltwo$ normalized
and afterwards aggregated. Note, due to the $\ltwo$ normalization, 
the actual linear PCA whitening cannot be applied after the aggregation.

In contrast to power normalization, the authors~\cite{Jegou14TEA} suggested the
use of \emph{rotation normalization} where the aggregated embedding is first
rotated by means of a PCA and then power-normalized. 

\subsection{Exemplar Support Vector Machines}
Recently, we proposed the use of Exemplar SVMs (ESVM) in the context of writer
recognition~\cite{Christlein17PR}. Since 
the training and test sets of common writer recognition datasets are disjoint,
an end-to-end classification of the writers is not possible.  However, this allows to 
compute probe individual classifiers. The encoding of each probe sample serves
as single positive instance while all samples from the training set are the 
negative samples for a linear support vector machine. This can be interpreted as computing
an individual similarity for each query document.
For evaluation, each other document is ranked according to the score of the
probe ESVM.

\section{Evaluation}
\label{sec:evaluation}
\subsection{Datasets}
We evaluate the following publicly available datasets:
\begin{itemize}
	\item 
The \icdard dataset~\cite{Louloudis13ICO} consists of four samples per
writer, where each writer contributed two samples in English and two samples in
Greek. The test set consists of
\num{250} writers and the training set contains
\num{100} writers.

\item The \cvl dataset~\cite{Kleber13CDA} (v.\,1.1) contains \num{310} writers.
\num{283} writers copied five texts (one in German and four in English). The
remaining \num{27} writers contributed seven texts
(one in German and six in English).
We use the first five forms of all \num{310} writers as test set and all 
samples of the IAM dataset~\cite{Marti02IAM} for training.

\item The \khatt dataset~\cite{Mahmoud14KOA} 
consists of \num{1000} writers, each contributing four documents written in Arabic. 
The dataset is
divided into three disjoint, \ie writer independent, sets for training
(\SI{70}{\percent}), validation (\SI{15}{\percent}) and testing
(\SI{15}{\percent}). 
\end{itemize}

The \cvl and \icdard datasets are text dependent, \ie each writer copied the same four to seven forms. 
Conversely, two of the four forms of the \khatt dataset contain unique paragraphs.
We binarized the \iam, \cvl, and \khatt datasets using Otsu's method~\cite{Otsu79}
to be more similar to the \icdard dataset and making them more
independent of the pen used.

The datasets are evaluated using a leave-one-sample-out cross validation. 
All results are given in terms of mean average precision
(\map), which computes the mean across the average precisions of all queries. 
The average precision of a query is
computed by ranking the remaining documents according to their similarity to the
query. Note all values are shown in percent. 

\subsection{CNN Activation Features}
We tested several architectures. First, we employ a four layer (not counting
pooling layers) deep CNN denoted as LeNet, which we used in our previous
work~\cite{Christlein15GCPR}. 
It has two convolutional layers, each followed by a max pooling layer, and
a fully connected layer followed by the classification layer. 
\begin{figure}[t]
	\tikzsetnextfilename{cnn_af_error}
	\begin{tikzpicture}
		\begin{axis}[scale only axis, 
				height=3.5cm, width=0.75\linewidth,
				xlabel={Epochs},
				ylabel={Error [\%]},
				xmin=1,xmax=200,
				xtick={1,20,...,200},
				xticklabels={0,1,...,10},
				ymin=38,ymax=100,
				ytick={40,50,...,100},
				no markers,
				legend cell align={left},
			]
			\addplot+[thick]table[x expr=\coordindex, y=test, y expr={100-\thisrow{test}}] 
			{icdar13_lenet_e25_3gr_fb_test.log};
			\addplot+[thick]table[x expr=\coordindex, y=test, y expr={100-\thisrow{test}}] 
			{icdar13_N1_e25_3gr_fb_test.log};
			\addplot+[thick,orange]table[x expr=\coordindex, y=test, y expr={100-\thisrow{test}}] 
			{icdar13_N3_e10_3gr_fb_test.log};
			\addlegendentry{LeNet (SGD)}
			\addlegendentry{ResNet-8 (SGD)}
			\addlegendentry{ResNet-22 (SGD)}
		\end{axis}
	\end{tikzpicture}
		\caption{Test error of the different architectures
	using the \icdard training data (subset).}
	\label{fig:test_error}
\end{figure}
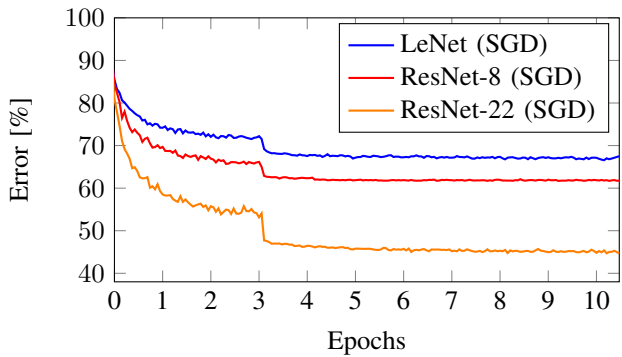

In addition to the LeNet architecture, we evaluate two different
ResNet models~\cite{He16a,He16b} of different depths.  We follow
the architectural design and training procedure of \etal{He}~\cite{He16a} for
the CIFAR10/100 datasets, where we evaluated eight and twenty layers deep CNNs,
further denoted as ResNet-8 and ResNet-20. 

The networks are optimized using SGD w.\,r.\,t.\ the cross-entropy loss, a
Nesterov momentum of $0.9$ and a weight decay of $10^{-4}$. We use an initial
learning rate of $0.01$ which is multiplied by $0.1$ after three epochs and once
more after four epochs.  The learning curves of the validation set (20\,k
independent patches) for the first ten epochs can be
seen in \cref{fig:test_error}.

Comparing the LeNet architecture with residual networks, 
the plot clearly shows the advantage of the latter. The deeper ResNet
reduced the error rate below \SI{50}{\percent}. This means in more than
\SI{50}{\percent} a $32\times32$ patch can be assigned to the correct writer.
However note that a deeper network also needs a longer training time.

\subsection{Decorrelation of CNN Activations}

\begin{table}[t]
	\centering
	\caption{Evaluation of different decorrelation methods for CNN activation
features extracted from the script contour (\map, avg.\ of 5 runs, \icdard
test set).}
\label{tab:local_norm_cnn}
	\begin{tabular}{lccc}
		\toprule
		Method & Baseline & PCA wh. &		ZCA wh. \\
		\midrule
		LeNet-A 	& 86.75 &  88.50 & 87.21 \\
		LeNet-B  	& 88.02 &  88.58 & 89.21 \\
		ResNet-8  & 88.39 &  89.83 & 88.68 \\
		ResNet-20 & 89.86 & \textbf{90.01} & 89.96 \\
		\bottomrule
  \end{tabular}
\end{table}

Given the trained models, we can use them as feature extractors by forwarding
the contour patches of a query sample through the networks. In all cases we used
the $64$-dimensional activations of the penultimate layer as features. They are
subsequently $\ltwo$-normalized.  For LeNet, we compared two different variants:
LeNet-A refers to the results using the very same LeNet model of our previous
work~\cite{Christlein15GCPR}, \ie the model after \num{20} epochs
trained by a fixed learning rate of $0.01$, while LeNet-B
denotes the results using the LeNet model after five epochs using the proposed
learning rate schedule. We also used the models after the fifth epoch for the
ResNet models since the error stagnated at this point.
Additionally, we evaluated the effect of decorrelating the activations by means
of i) PCA rotation and whitening and ii) zero component analysis (ZCA) with
whitening. ZCA whitening rotates the data back to be as close to the input as
possible. In the field of deep learning, ZCA whitening is more common than PCA
whitening.

For the baseline encoding, we employ VLAD encoding (a pure sum pooling, \ie no
embedding achieves 78.80 \map).
Since VLAD encoding depends much on a good background model, we compute five different
VLAD-encodings and give the average result. The background model is computed by
a mini-batch version of $k$-means~\cite{Sculley10WSK} using different seeds and
different selection of training samples. In this way, we have a large
variability in the background models. The VLAD encodings are individually
normalized by power normalization ($p=0.5$) followed by an
$\ltwo$ normalization. 

\Cref{tab:local_norm_cnn} depicts all \map results. The baseline results (first
column) show 
that LeNet-A is inferior to LeNet-B, suggesting that a learning rate schedule is
beneficial for the final feature representation. Any LeNet configuration
performs worse than ResNets. A decorrelation is especially beneficial for the
smaller networks. However, no decorrelation method is significantly superior. 
Interestingly, a PCA whitening of the ResNet-20 model achieves the highest
results. 
Exchanging SSR with intra
normalization~\cite{Arandjelovic13AAV} or residual
normalization~\cite{Delhumeau13} for the baseline experiment performed worse with 89.44 and 88.31 \map,
respectively (no normalization at all: 88.46 \map).
In the following experiments, we used the activation features of the
ResNet-20 model, if not mentioned otherwise in its non-whitened version.

\subsection{From VLAD to Triangulation Embedding}
\begin{table}[t]
	\caption{Comparison of VLAD in combination with LCS with triangulation
		embedding (avg.\ of 5 runs, \icdard test set).
}
\label{tab:vladplus}
	\centering
	\begin{tabular}{lc}
	\toprule
	Method & \map \\
	\midrule
	VLAD + ResNorm + LCS wh. & \textbf{90.44}\\
	VLAD + ResNorm + LCS++ & 90.42\\
	T-Emb & 89.71\\
	T-Emb16 & 87.65\\
	\bottomrule
\end{tabular}
\end{table}

Since triangulation embedding is an adapted VLAD encoding, it is interesting to explore
the individual adjustments. Instead of T-emb's global PCA whitening, we use
LCS~\cite{Delhumeau13} with whitening on $\ltwo$ normalized residuals (VLAD + ResNorm + LCS wh.). This is
compared with a variant where the first component of each PCA transformed 
$\vec{\phi}_k$ is discarded (VLAD + ResNorm + LCS++). 
Two variants of triangulation embedding are used for comparison. The
first one uses $100$ cluster centers similar to VLAD (T-Emb), and the other one uses $16$
components (T-Emb16). The latter is the default value of triangulation
embedding~\cite{Jegou14TEA} to speed up the aggregation process.
\Cref{tab:vladplus} shows that VLAD in combination with LCS whitening is
superior to triangulation embedding. This suggests that the power of
triangulation is not based on the aggregation of more descriptors but instead on
the positive effect of PCA whitening.

To make the evaluation process completely fair, we also evaluated the effect of
rotation normalization (RotNorm), which was proposed in combination with triangulation
embedding~\cite{Jegou14TEA}. Additionally, we evaluated a simple PCA whitening
(PCA wh.) and a joint PCA whitening (joint PCA wh.), where the encodings of the
five runs are concatenated and jointly whitened.
\Cref{tab:temb_af} shows that another decorrelation of the final aggregated
embedding is not beneficial. While a joint PCA whitening shows some improvements
for triangulation embedding, it worsens the results of VLAD plus residual
normalization and LCS whitening (VLAD++).
\begin{table}[t]
	\centering
	\caption{Evaluation of CNN activation features in combination with T-Emb
	(\map, avg.\ of 5 runs, \icdard test set).} 
	\label{tab:temb_af}
	\begin{tabular}{l*{4}c}
			\toprule
			Method & VLAD++ & T-Emb & T-Emb16 \\
			\midrule
			Baseline &\textbf{90.44} & 89.71  & 87.65\\
			RotNorm & 81.73 & 88.78 & 88.81\\
			PCA wh.\ & 82.23 & 89.39 & 89.60\\
			joint PCA wh.\ & 83.04 & 90.08 & 90.21\\
		\bottomrule
	\end{tabular}
\end{table}

\subsection{Sum Pooling vs.\ Generalized Max Pooling}
The use of GMP requires the proper choice of the regularization parameter
$\lambda$. \etal{Murray}~\cite{Murray16} suggested that a factor of $1.0$
works well in practice. However, cross-validating its effect on the \icdard
training set showed that
a much higher value of $\lambda=1000$ is preferable among the embeddings. 

In \cref{tab:gmp}, we compare sum pooling with GMP using the two embeddings
VLAD++ and T-Emb with $16$ components (T-Emb16).
When using a proper regularization parameter, GMP outperforms sum pooling by a
small factor. However, the benefit is small and another parameter ($\lambda$)
needs to be cross-validated. 
\begin{table}[t]
	\centering
	\caption{Comparison of sum pooling and GMP (\map, avg.\ of 5 runs, \icdard
	test set).} 
	\label{tab:gmp}
	\begin{tabular}{lccc}
		\toprule 
		Method & Sum & GMP$_{\lambda=1}$ & GMP$_{\lambda=10^3}$\\
		\midrule
		VLAD++ 	& 90.44 & 89.74 & \textbf{90.66}\\
	T-Emb16 											& 87.65 & 78.59 & \textbf{89.11}\\
		\bottomrule
	\end{tabular}
\end{table}

\subsection{Exemplar Support Vector Machines and PCA whitening}
We compare the effectiveness of ESVMs with: VLAD plus SSR (VLAD), VLAD with LCS
whitening (VLAD++) and triangulation embedding (T-Emb16). All methods use GMP
($\lambda=1000)$, the margin parameter of the SVM is for each method
cross-validated using the training set.  Comparing the different results of
\cref{tab:esvm}, we see that all methods perform well.  Interestingly, the basic
VLAD method performs the best. Hence, we can conclude
that the additional decorrelation with PCA, which is part of LCS and
triangulation embedding does not necessarily improve the recognition results. 
\begin{table}[t]
	\centering
	\caption{Evaluation of ESVMs using (a) different embeddings, aggregated with
		GMP and (b) additional PCA whitening of the CNN activation features
		(\map, avg.\ of 5 runs, \icdard test set).}
	\label{tab:esvm_full}

	\subcaptionbox{CNN-AF\label{tab:esvm}}{
	\begin{tabular}{lcc}
		\toprule 
		Method & Baseline & ESVM\\
		\midrule
		VLAD 		& 89.86 & \textbf{91.72}\\  
		VLAD++ 	& 90.66 & 91.64\\
		T-Emb16 & 89.11 & 91.56\\
		\bottomrule
	\end{tabular}
}
\subcaptionbox{CNN-AF + PCA wh.\label{tab:esvm_pwh}}[0.45\linewidth]{
	\begin{tabular}{cc}
		\toprule 
		Baseline & ESVM\\
		\midrule
		 	90.19 & \textbf{93.24}\\
		 	90.41 & 91.53\\
		 	89.24 & 92.73\\
		\bottomrule
	\end{tabular}
}
\end{table}
Additionally, we investigated the effect of the PCA whitened version of 
the CNN activation features.  \Cref{tab:esvm_pwh} shows that this step improves
the recognition of the basic VLAD method and triangulation embedding.

\subsection{Comparison with the State of the Art}
Finally, we compare the results with the state of the art on the \icdard, \cvl,
and \khatt datasets.
Additional to the \map, we depict the following common writer
identification\,/\,retrieval measures:
Top-1 gives the probability that the first retrieved result stems from the same
writer. While Soft-5 (S-5)\,/\,Soft-10 (S-10) requires that one
of the top five\,/\,ten results is written by the query writer, 
Hard-2 (H-2)\,/\,Hard-3 (H-3) requires that all
two\,/\,three top ranked results are from the same writer.

We used VLAD in combination with generalized max pooling
($\lambda=1000$) and SSR. Additionally, we show the results with the Exemplar
SVM step. We compare our results with our previous work~\cite{Christlein17PR}
and the work of Tan and Wu~\cite{Tang16}. To the best of our knowledge, these
two methods are currently leading the \icdard, \cvl and \khatt datasets.

\Cref{tab:sota} shows all results. Our method achieves results similar to the 
state of the art on all datasets while even being superior on the \icdard and \khatt
dataset. Previous evaluations on the \cvl dataset~\cite{Christlein17PR} suggest
that a better background model could further improve the results. 

\begin{table}[t]
	\centering
	\caption{Comparison with the state of the art.}
	\label{tab:sota}
	\subcaptionbox{\icdard}
	{
		\begin{tabular}{l*{7}cc}		
		\toprule
		Method & Top-1 & H-2 & H-3 & S-5 & S-10 & \map\\ 
		\midrule
		\cite{Tang16} & 99.0 & 84.4 & 68.1 & 99.2 & 99.6 & -- \\
		\cite{Christlein17PR} & \textbf{99.7} & 84.8 & 63.5 &
		\textbf{99.8} & 99.8 & 89.4\\
		\midrule
		VLAD 				& 99.0 & 85.3 & 68.6 & 99.4 & 99.7 & 90.2\\
		VLAD + E & 99.6 & \textbf{89.8} & \textbf{77.0} & \textbf{99.8} &
		\textbf{99.9} & \textbf{93.2}\\
		\bottomrule
	\end{tabular}
}	

\medskip
\subcaptionbox{\cvl}{
		\begin{tabular}{l*{7}cc}		
		\toprule
		Method & Top-1 & H-2 & H-3 & S-5 & S-10 & \map\\ 
		\midrule
		\cite{Tang16} & \textbf{99.7} & \textbf{99.0} & \textbf{97.9} &
		\textbf{99.8} & \textbf{100} & --\\
		\cite{Christlein17PR} & 99.2 & 98.4 & 97.1 & 99.6 & 99.7 
		& 98.0\\
		\midrule
		VLAD 				 & 99.2 & 98.4 & 96.1 & 99.5 & 99.6 &	97.4\\
		VLAD + E & 99.5 & \textbf{99.0} & 97.7 & 99.6 & 99.8 &	\textbf{98.4}\\
		\bottomrule

	\end{tabular}
	}

	\medskip
	\subcaptionbox{\khatt}{
		\begin{tabular}{l*{7}cc}		
		\toprule
		Method & Top-1 & H-2 & H-3 & S-5 & S-10 & \map\\ 
		\midrule
		
		\cite{Christlein17PR} & 99.5 & 96.5 & 92.5 &  99.5 & 99.5&
		97.2\\
		\midrule
		VLAD 					& 99.1 & 93.8 & 88.2 & 99.4 & 99.6 &  95.5\\
		VLAD + E 	& \textbf{99.6} & \textbf{97.6} & \textbf{94.5} &
		\textbf{99.7} & \textbf{99.7} & 
		\textbf{98.0}\\

		\bottomrule
	\end{tabular}
	}
\end{table}

\section{Conclusion}
\label{sec:conclusion}
In this work, we evaluated several effects related to the encoding of local
descriptors. We showed that 1) a deeper CNN only slightly achieves higher
retrieval performance 2) triangulation embedding is not better than VLAD when
either the local descriptors or the embeddings are 
decorrelated by means of a PCA whitening. 3) Generalized max pooling is slightly
better than sum pooling but is sensitive to a proper regularization. 4) When using Exemplar SVMs,
all three methods perform similarly well, where VLAD in combination with power
normalization performs slightly better. The
proposed combination of deep CNN activation features, VLAD encoding,
normalization and Exemplar SVMs is similar or better than state-of-the-art
methods on three publicly available datasets.  
For future work, we would like to explore different layers than the penultimate
layer for feature extraction. Furthermore, larger input patch sizes could also
be beneficial.

\bibliographystyle{IEEEtran}
{\small
\bibliography{das}
}
\end{document}